\documentclass[conference,a4paper]{IEEEtran}
\IEEEoverridecommandlockouts

\usepackage[hidelinks]{hyperref}
\usepackage[cmex10]{amsmath}
\usepackage{amssymb,amsfonts}
\usepackage{dblfloatfix}

\usepackage{float}
\usepackage[ruled,vlined]{algorithm2e}
\usepackage{graphicx}
\graphicspath{{Figures/PDF/}{Figures/PNG/}}

\usepackage{booktabs}
\usepackage{siunitx}
\usepackage[numbers,compress]{natbib}
\usepackage{texnames}
\usepackage{bm,bbm}
\usepackage{orcidlink}

\newcommand{\figcaption}[1]{\par\vspace{-10pt}\caption{#1}\vspace{-2pt}}
\AtBeginDocument{%
	\setlength{\textfloatsep}{8pt}%
	\setlength{\dbltextfloatsep}{8pt}%
}

\begin{document}

\title{Orbit-Planner: Towards Latent World Models for On-Orbit Obstacle Avoidance of Satellite Agents
\thanks{This work was supported by the National Key R\&D Program of China under Grant 2024YFF1401001, and partially supported by China Science and Technology Cloud (CSTCloud).
    
    Zhijian Li, Chao Ren, Peijin Wang and Xian Sun are with the Aerospace Information Research Institute, Chinese Academy of Sciences, Beijing 100190, China, also with the School of Electronic, Electrical and Communication Engineering, University of Chinese Academy of Sciences, Beijing 100190, China, also with the University of Chinese Academy of Sciences, Beijing 100190, China, and also with the Key Laboratory of Target Cognition and Application Technology (TCAT), Aerospace Information Research Institute, Chinese Academy of Sciences, Beijing 100094, China (e-mail: lizhijian25@mails.ucas.ac.cn, renc0003@e.ntu.edu.sg, wangpeijin17@mails.ucas.ac.cn, sunxian@aircas.ac.cn).
    \IEEEauthorrefmark{1}Corresponding author: Chao Ren.}}

\author{	\IEEEauthorblockN{Zhijian Li\orcidlink{0009-0009-9067-8862}, Chao Ren\IEEEauthorrefmark{1}\orcidlink{0000-0001-9096-8792}, Peijin Wang\orcidlink{0000-0002-0371-5584}, Xian Sun\orcidlink{0000-0002-0038-9816}~\IEEEmembership{Senior Member,~IEEE}}
	\IEEEauthorblockA{\textit{Aerospace Information Research Institute, Chinese Academy of Sciences, 100094 Beijing, China}\\
		lizhijian25@mails.ucas.ac.cn, renc0003@e.ntu.edu.sg\\[4pt]
		\includegraphics[height=1.2em]{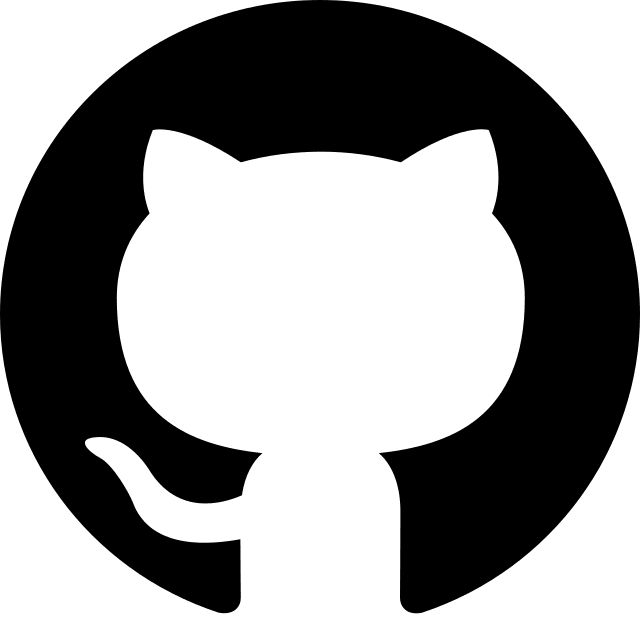}\hspace{2pt}\href{https://github.com/ZhijianLi2003/Orbit_Planner}{\textcolor{blue}{Code}}\hspace{12pt}%
		\includegraphics[height=1.2em]{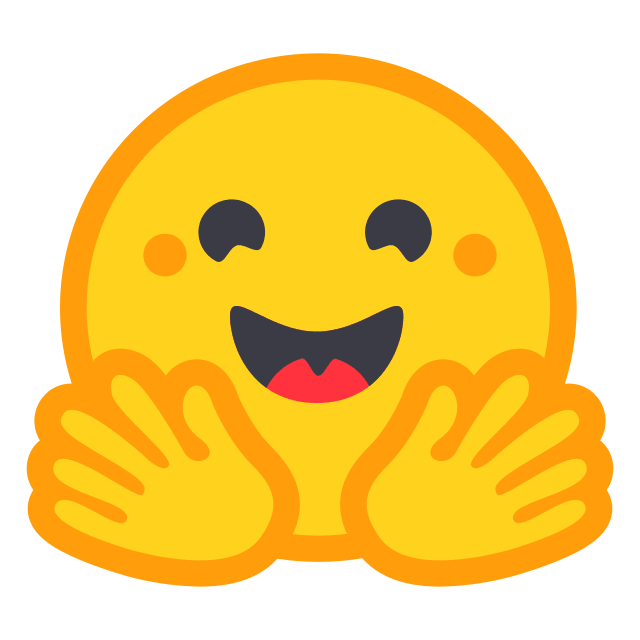}\hspace{2pt}\href{https://huggingface.co/datasets/warriorLZJ/Orbit_Planner}{\textcolor{blue}{HuggingFace}}\hspace{12pt}%
		\href{https://zhijianli2003.github.io/Orbit_Planner/}{\textcolor{blue}{Project Page}}}
}

\maketitle
\begin{abstract}
	Satellite agents for on-orbit navigation tasks need to predict collision risks using limited onboard observations. However, conventional planners often rely on predefined maps and fixed environmental assumptions, limiting their adaptability in dynamic on-orbit scenarios. In this paper, we propose Orbit-Planner, a two-stage latent world model for on-orbit obstacle avoidance. Orbit-Planner learns action-conditioned spacecraft dynamics to perform future-state rollouts in latent space, and introduces a Physics Probe to decode physical state changes from imagined latent trajectories. Experiments demonstrate that Orbit-Planner can perform long-horizon latent rollouts and recover physical states from imagined trajectories. In closed-loop obstacle-avoidance navigation in Isaac Sim, it attains a success rate of $\mathbf{91.7\%}$. Code is available at \url{https://github.com/ZhijianLi2003/Orbit_Planner}.
\end{abstract}

\begin{IEEEkeywords}
	Latent world models, Space robotics, Autonomous navigation, On-orbit obstacle avoidance
\end{IEEEkeywords}

\section{Introduction}

With the rapid increase in space exploration and on-orbit assembly tasks, autonomous satellite agents are playing an increasingly critical role~\cite{orsula2025space}. In these complex and dynamic orbital environments, effective obstacle avoidance is paramount to ensure the safety and success of space missions. Traditional methods often rely on predefined environmental parameters and classical path-planning algorithms, such as Artificial Potential Fields and A* search~\cite{li2025obstacle}. Imitation-learning policies~\cite{chi2025diffusion} can generate reactive maneuvers from demonstrations, but they often generalize poorly when the test-time dynamics deviate from the demonstrated distribution. 

To overcome these issues, world models have emerged as a transformative paradigm. By extracting compact predictive representations from high-dimensional sensory inputs, these models empower agents to internally simulate future states and evaluate actions prior to physical execution~\cite{hafner2025mastering, hansen2024td}. Notably, recent breakthroughs in Joint-Embedding Predictive Architectures (JEPA) have enabled stable, end-to-end learning of such models~\cite{maes2026leworldmodel}, demonstrating remarkable efficacy in highly dynamic control scenarios like agile quadrotor flight~\cite{rao2026skyjepa}. Inspired by these advancements, latent world models present a compelling pathway for space robotics. 

\begin{figure}[!t]
	\centering
	\includegraphics[width=0.88\columnwidth]{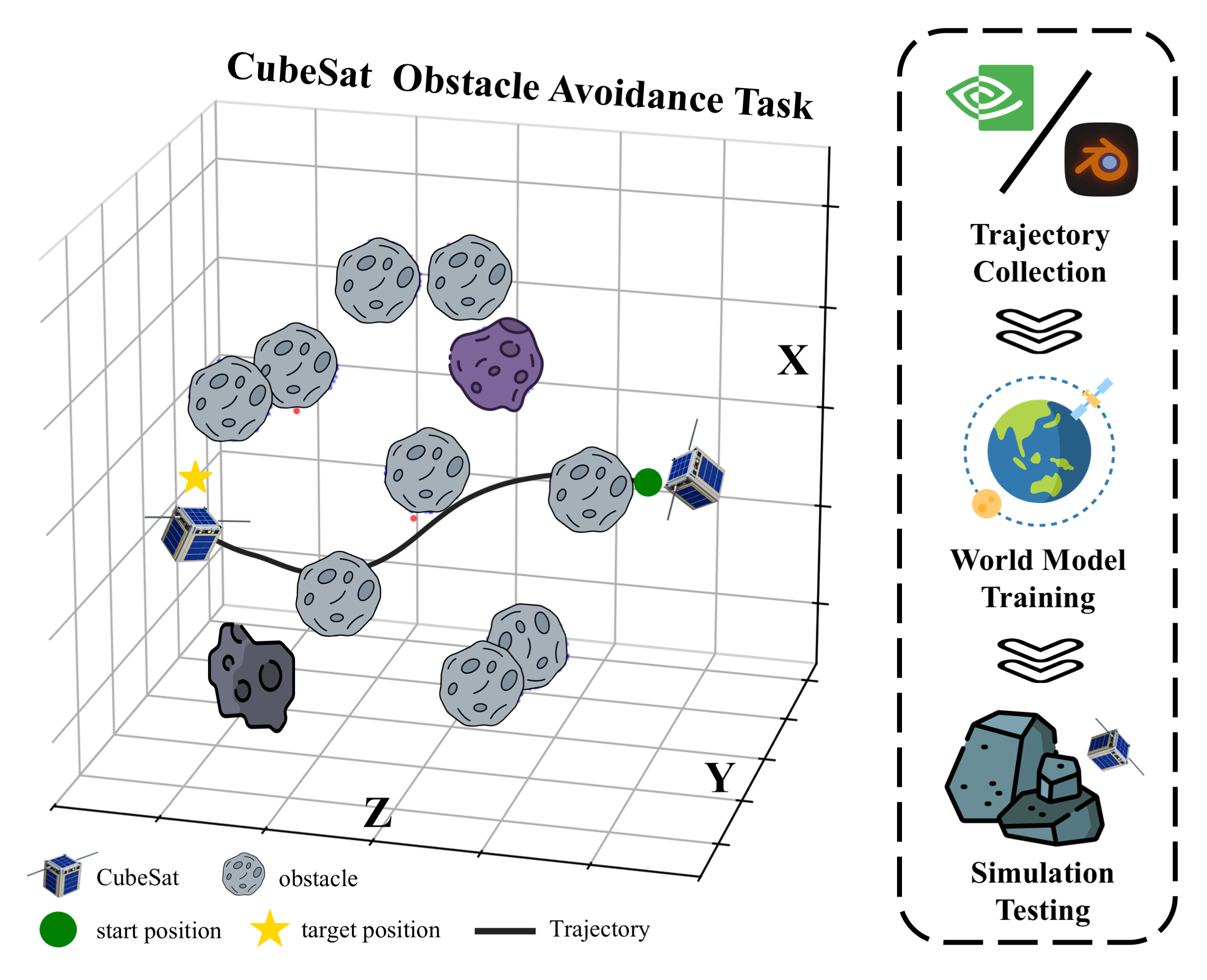}
	\figcaption{Overview of the on-orbit obstacle-avoidance task and the overall pipeline. Left: a CubeSat navigates from a start position to a target through a cluttered orbital field of obstacles. Right: the pipeline of data collection, world-model training, and testing.}
	\label{fig:overview}
\end{figure}

The main contributions of this paper are as follows: 1) we construct an on-orbit obstacle-avoidance dataset for satellite agents based on Isaac Sim; 2) we propose Orbit-Planner, a two-stage latent world model that learns action-conditioned dynamics for future-state rollout, with a Physics Probe to recover physical states from imagined trajectories; 3) based on this world model, we achieve online on-orbit obstacle-avoidance navigation, with a $91.7\%$ success rate.

\section{Methodology}

\begin{figure*}[!t]
	\centering
	\includegraphics[width=\textwidth]{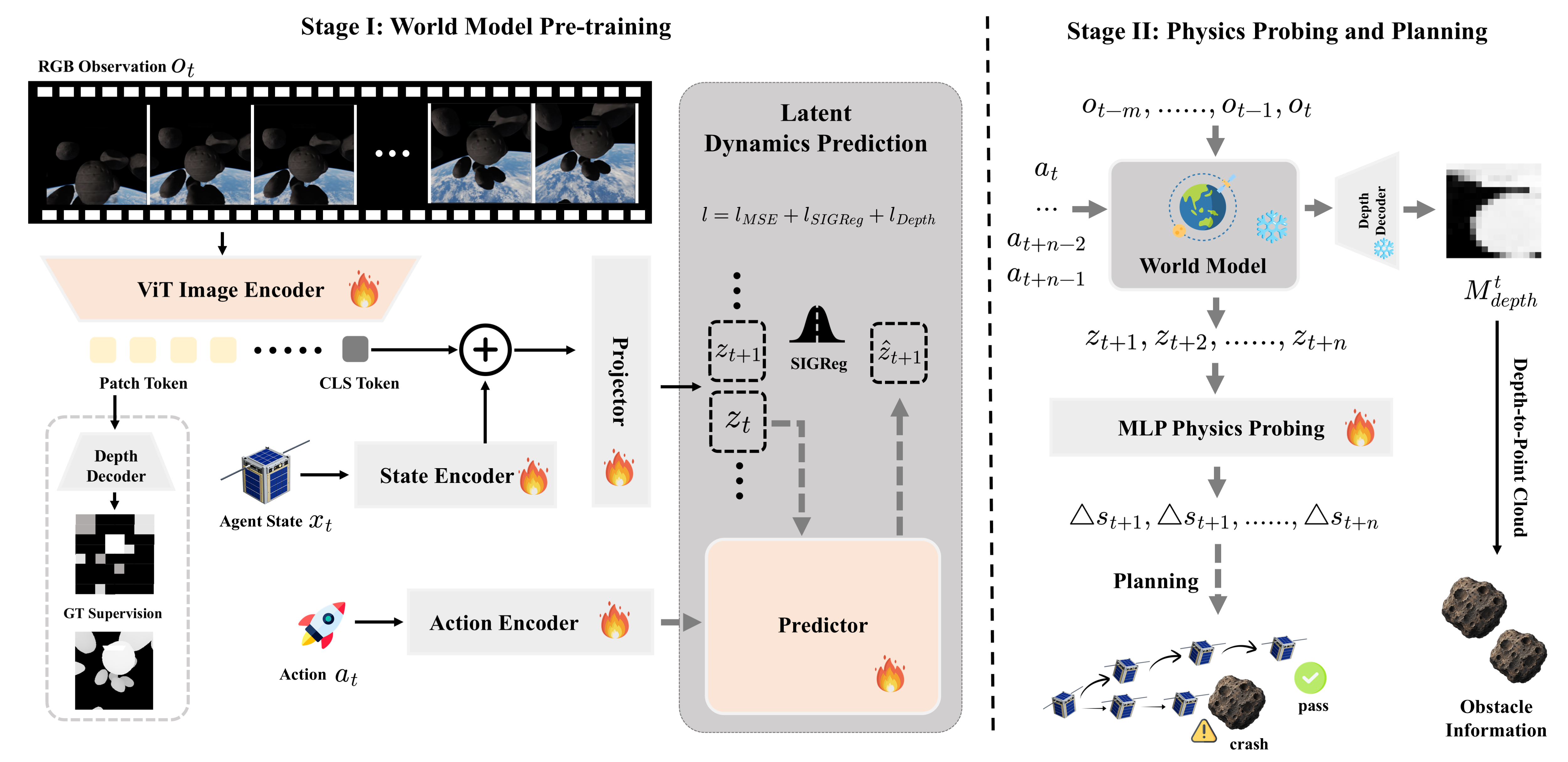}
	\figcaption{Overview of the proposed Orbit-Planner framework. Stage I pre-trains the latent world model to represent multimodal observations (RGB sequences and spacecraft states) and perform action-conditioned latent rollouts. In Stage II, the Physics Probe maps rolled-out latents to future spacecraft-state increments, while the depth decoder recovers current obstacle geometry; both are used for trajectory planning.}\label{fig:framework}
\end{figure*}

\subsection{Problem Formulation}
We consider a discrete-time dynamical system for a CubeSat agent operating in a dynamic and cluttered orbital environment. As illustrated in Fig.~\ref{fig:overview}, the primary objective is to navigate the satellite safely to a target destination while minimizing the probability of collision. The system is characterized by its state $\bm{s}_t \in \mathbb{R}^{16}$ and control action $\bm{a}_t \in \mathbb{R}^8$.

For the CubeSat, we define the state as:
\begin{equation}
    \bm{s}_t = [\bm{p}_t^\top \quad \bm{v}_t^\top \quad \bm{r}_t^\top \quad \bm{\omega}_t^\top \quad \phi_t]^\top.
\end{equation}
Here, $\bm{p}_t \in \mathbb{R}^3$ denotes the position displacement in the inertial (world) frame relative to the initial position ($\bm{p}_0 = \bm{0}$). The linear velocity $\bm{v}_t \in \mathbb{R}^3$ and angular velocity $\bm{\omega}_t \in \mathbb{R}^3$ are expressed in the body frame. The attitude is represented by a continuous 6D rotation vector $\bm{r}_t \in \mathbb{R}^6$, derived from the first two columns of the rotation matrix $\bm{R}_t \in SO(3)$ representing the attitude of the body frame in the inertial frame. Additionally, $\phi_t \in [0, 1]$ represents the remaining fuel ratio. The discrete time step is set to $\Delta t = 0.04$\,s (\SI{25}{\hertz}).

The control action $\bm{a}_t \in \mathbb{R}^8$ denotes the normalized thrust commands for the 8 onboard thrusters:
\begin{equation}
    \bm{a}_t = [F_{1,t}, F_{2,t}, \dots, F_{8,t}]^\top \in [0, 1]^8.
\end{equation}

The system operates under the assumption that the agent's physical state $\bm{s}_t$ is fully observable. To capture surrounding environmental information, the agent utilizes an onboard RGB camera to obtain visual observations $\bm{o}_t \in \mathbb{R}^{H \times W \times 3}$.

\subsection{Orbit-Planner World Model Framework}
As illustrated in Fig.~\ref{fig:framework}, Orbit-Planner consists of two stages: world-model pre-training, followed by physics probing and planning.

\subsubsection{Stage I: World Model Pre-training}
In the first stage, we pre-train a latent world model to capture the complex dynamics of the space environment. This stage fundamentally consists of two processes: representation and rollout. During the representation phase, the model takes RGB observations $\bm{o}_t$ and low-dimensional spacecraft states $\bm{s}_t$ as inputs. A Vision Transformer (ViT) acts as the visual encoder, extracting a global class token and dense patch tokens. Concurrently, an MLP-based State Encoder maps the physical state into a state embedding. A Projector module then fuses the class token and the state embedding to generate the latent representation $\bm{z}_t$, which jointly encodes the agent's own physical state and the surrounding environmental geometry. Crucially, this projector employs Batch Normalization to counteract the LayerNorm effects from the ViT, ensuring that the variance-based regularization (SIGReg~\cite{maes2026leworldmodel}) functions effectively on the latent distribution without collapsing. The dense patch tokens are fed to a dedicated depth decoder for depth prediction.

During the rollout phase, the control action $\bm{a}_t$ is introduced. The thruster commands are processed by an MLP-based Action Encoder to produce an action embedding. The Latent Dynamics Prediction module, parameterized by an autoregressive Transformer equipped with Adaptive Layer Normalization (AdaLN), then utilizes the current latent state $\bm{z}_t$ and the action embedding to predict the future latent state $\bm{z}_{t+1}$. Specifically, the action embedding acts as the conditioning signal in the AdaLN blocks~\cite{peebles2023scalable}, dynamically modulating the latent features to ensure the physical actions effectively guide the future-state rollout. To ensure the learned representations are physically meaningful and geometrically consistent, the training is guided by a composite loss function $\mathcal{L}$:
\begin{equation}
    \mathcal{L} = \mathcal{L}_{\text{MSE}} + \mathcal{L}_{\text{SIGReg}} + \mathcal{L}_{\text{Depth}},
\end{equation}
which incorporates the mean squared error ($\mathcal{L}_{\text{MSE}}$) for latent dynamics prediction, variance-based regularization constraints ($\mathcal{L}_{\text{SIGReg}}$) to prevent feature collapse, and depth supervision ($\mathcal{L}_{\text{Depth}}$) via a dedicated depth decoder. The depth loss is included mainly to encourage the visual encoder to capture depth cues that are critical for navigation.

\subsubsection{Stage II: Physics Probing and Planning}
After pre-training, we freeze the world model and use it to roll out future states. Given a sequence of past observations $(\bm{o}_{t-m+1}, \dots, \bm{o}_t)$, the corresponding states $(\bm{s}_{t-m+1}, \dots, \bm{s}_t)$, and a sequence of proposed future actions $(\bm{a}_t, \dots, \bm{a}_{t+n-1})$, the world model predicts the corresponding future latent states $(\bm{z}_{t+1}, \dots, \bm{z}_{t+n})$. During training, the observation context length is set to $m=8$ and the rollout horizon is set to $n=12$. To map latent states to physical quantities, we introduce a Physics Probe and train only this probe while keeping the world model frozen. The probe translates the latent states into physical state transitions $(\Delta \bm{s}_{t+1}, \dots, \Delta \bm{s}_{t+n})$. Simultaneously, the frozen depth decoder extracts obstacle information, converting it into a Depth-to-Point Cloud representation.

\begin{figure}[!t]
	\centering
	\includegraphics[width=\columnwidth]{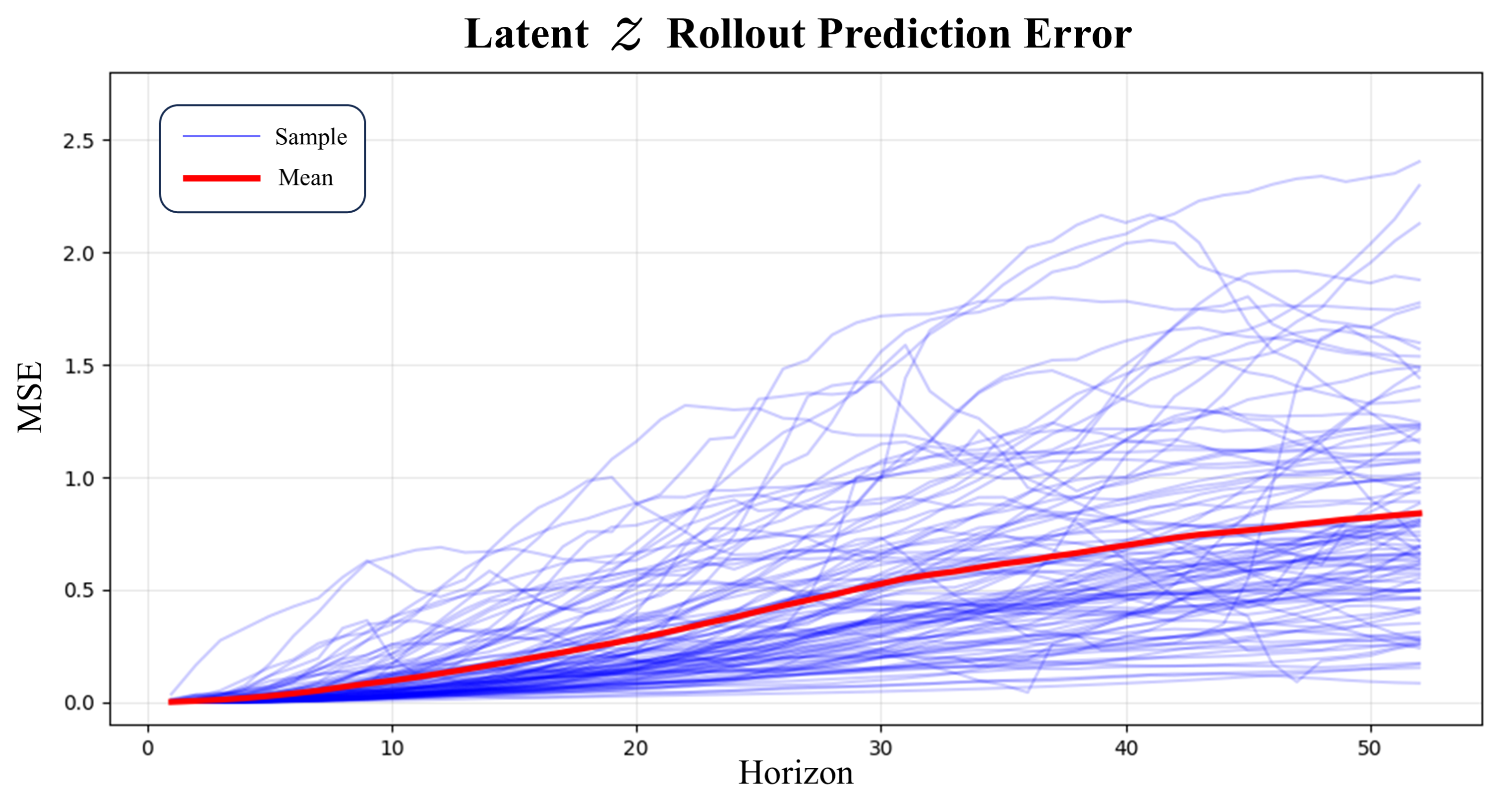}
	\figcaption{Latent-space rollout prediction error versus horizon. Thin blue curves denote individual trajectories; the red curve denotes the mean MSE.}
	\label{fig:latent_rollout}
\end{figure}

\subsection{Autonomous Control and Decision-Making}
Given the probe-predicted state rollouts and obstacle point clouds, Orbit-Planner employs MPPI~\cite{williams2015model} with $K=256$ samples and a horizon of $H=50$ steps, and selects the sequence with the lowest predicted collision risk while progressing toward the target. Only the first action of the selected sequence is executed before replanning.

\section{Experiments}

\subsection{Experimental Setup}
We collect data in a space robotics simulation environment~\cite{orsula2025space} based on Isaac Sim. The dataset contains 8000 on-orbit obstacle-avoidance trajectories, split into 7200 for training and 800 for testing. To improve the robustness of the learned representations under visual and geometric variations, extensive domain randomization is applied during data collection, encompassing variations in obstacle positions and lighting conditions. Furthermore, to improve behavioral diversity, trajectories are gathered under three distinct levels. Expert trajectories are generated by RRT$^*$ path planning followed by PD tracking, with larger safety margins and lower control noise. Risky trajectories fly straight toward the goal without obstacle-aware planning. Exploratory trajectories inject strong random action noise for unstructured exploration. 

\begin{figure}[!t]
	\centering
	\includegraphics[width=\columnwidth]{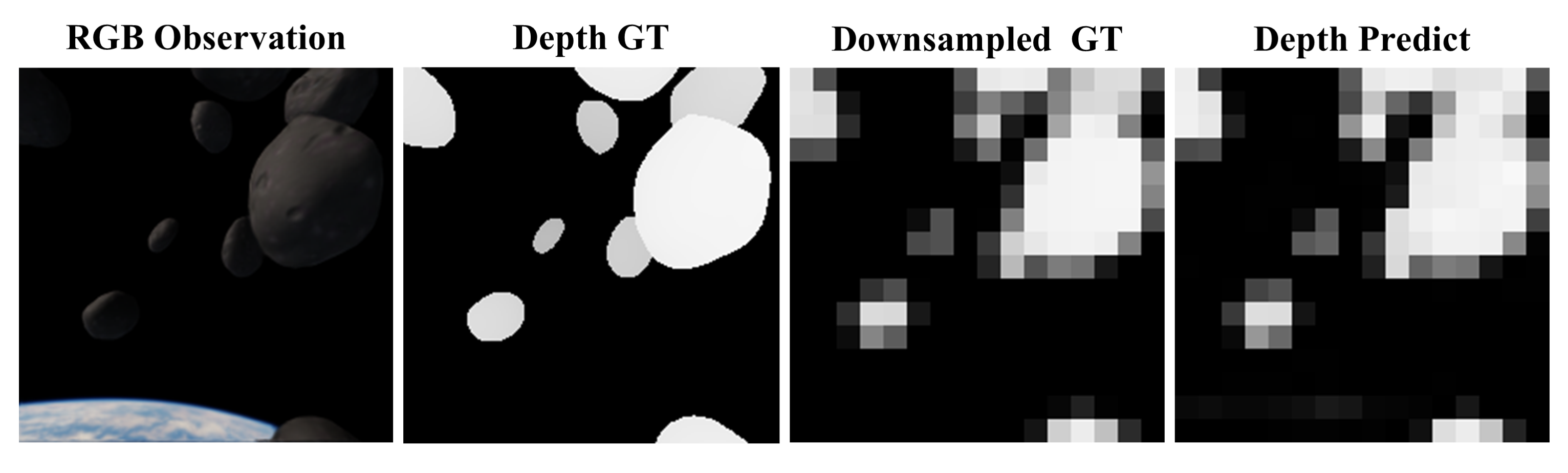}
	\figcaption{RGB-to-depth prediction results. Both the ground-truth and predicted depth maps are downsampled to $16\times16$.}
	\label{fig:depth_predict}
\end{figure}

\begin{table}[!t]
	\centering
	\caption{Physics Probe errors on absolute states recovered via kinematic integration from predicted increments $\Delta \bm{s}$ (mean $\pm$ std over 100 randomly selected testing trajectories). }
	\label{tab:physics_probe}
	{\scriptsize
	\setlength{\tabcolsep}{3pt}
	\renewcommand{\arraystretch}{0.92}
	\begin{tabular}{lcc}
	\toprule
	Quantity & \begin{tabular}[c]{@{}c@{}}MAE\\($h = 25, t = 1\text{s}$)\end{tabular} & \begin{tabular}[c]{@{}c@{}}MAE\\($h = 50, t = 2\text{s}$)\end{tabular} \\
	\midrule
	Position $\bm{p}$ (m)  & $0.0394 \pm 0.0554$ & $0.0813 \pm 0.0661$ \\
	Velocity $\bm{v}$ (m/s)  & $0.0636 \pm 0.0963$ & $0.1191 \pm 0.1102$ \\
	Rotation (geodesic, rad)  & $0.0676 \pm 0.0909$ & $0.2034 \pm 0.1826$ \\
	Angular velocity $\bm{\omega}$ (rad/s)  & $0.0883 \pm 0.1085$ & $0.1557 \pm 0.1453$ \\
	Fuel $\phi$ (normalized)  & $0.0019 \pm 0.0015$ & $0.0033 \pm 0.0021$ \\
	\bottomrule
	\end{tabular}}
	\end{table}

\subsection{Latent Space Rollout}

\begin{figure*}[!t]
	\centering
	\includegraphics[width=\textwidth]{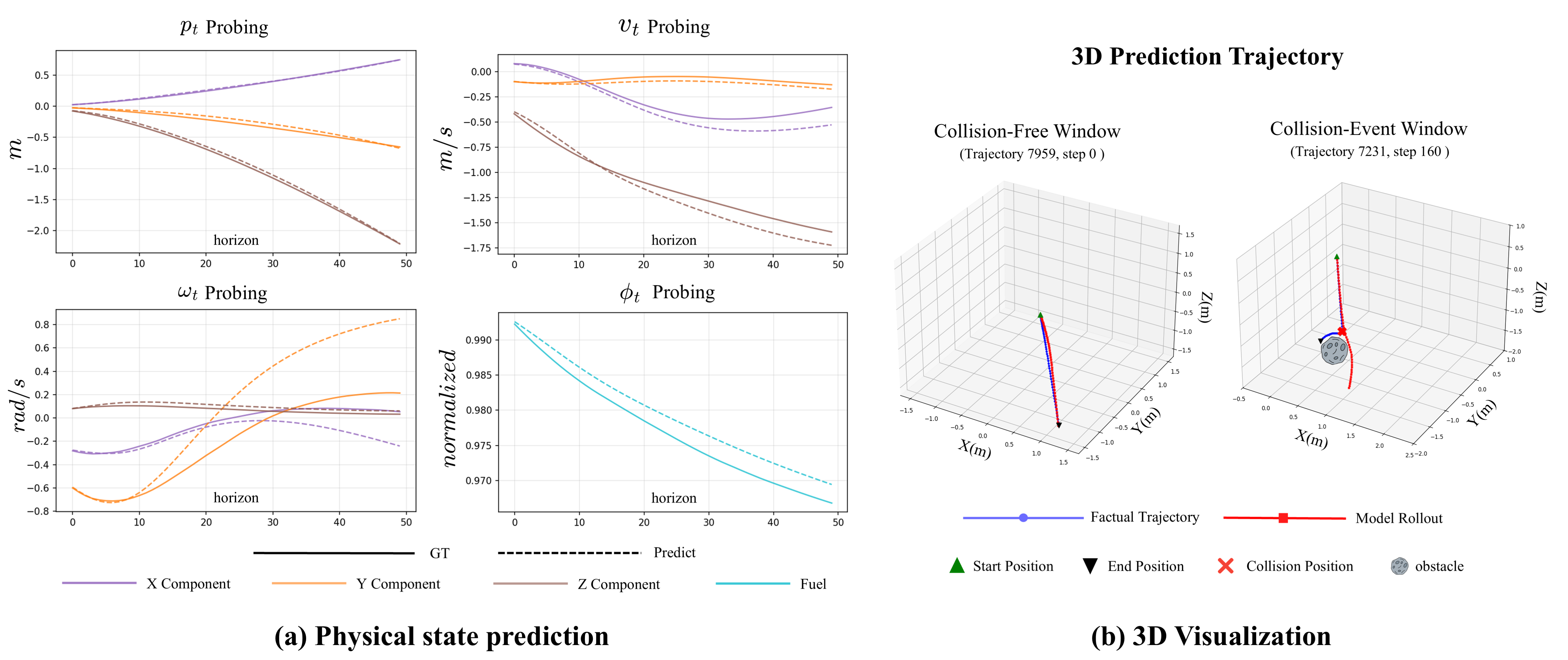}
	\figcaption{Physics Probe qualitative results on a 50-step (\SI{2.0}{\second}) prediction horizon. (a) Predicted versus ground-truth physical-state components on a collision-free trajectory. (b) 3D comparison of a collision-free window and a collision window.}
	\label{fig:physics_probe}
\end{figure*}

We evaluate action-conditioned multi-step prediction in the latent space $\bm{z}$. Starting from an encoded context, the frozen world model recursively rolls out future latents under the recorded action sequence, and we measure the MSE against encoder-derived target latents at each horizon. Fig.~\ref{fig:latent_rollout} shows the per-trajectory errors (blue) and their mean (red). As expected for autoregressive imagination, the error accumulates with the horizon; nevertheless, the mean MSE grows gradually and remains moderate even at a 50-step horizon, indicating that the learned latent dynamics support stable long-horizon rollouts for action-conditioned trajectory imagination. Fig.~\ref{fig:depth_predict} shows that the depth decoder recovers obstacle structure, indicating that $\bm{z}$ preserves useful geometric cues.

\subsection{Physics Probe}

We evaluate the accuracy of physical-state recovery from action-conditioned latent rollouts. With the world model's encoder and predictor frozen, an 8-step observation context is encoded into initial latent states. The model then autoregressively rolls out future latents conditioned on the recorded action sequences. Subsequently, the Physics Probe maps each predicted latent state to physical state increments $\Delta \bm{s}$, which are kinematically integrated from the last context state to reconstruct the full physical trajectory. We compare these recovered states against the ground truth and report the Mean Absolute Error (MAE) across 25-step and 50-step prediction horizons. As shown in Table~\ref{tab:physics_probe}, position, velocity, and fuel are recovered with low error, whereas angular velocity exhibits slightly higher variance due to its highly dynamic nature. Fig.~\ref{fig:physics_probe} provides a qualitative comparison. Within a limited time horizon, the collision-free rollout remains aligned with the ground-truth trajectory. In a prediction window that contains a collision event, the error between the rollout and the recorded trajectory becomes substantially larger, which indirectly suggests that the model has captured action-conditioned dynamics rather than merely replaying the observed outcome. These results indicate that the action-conditioned latent rollouts preserve physically meaningful dynamics that can be decoded by the probe.

\subsection{On-Orbit Obstacle Avoidance}

\begin{figure}[!t]
	\centering
	\includegraphics[width=\columnwidth]{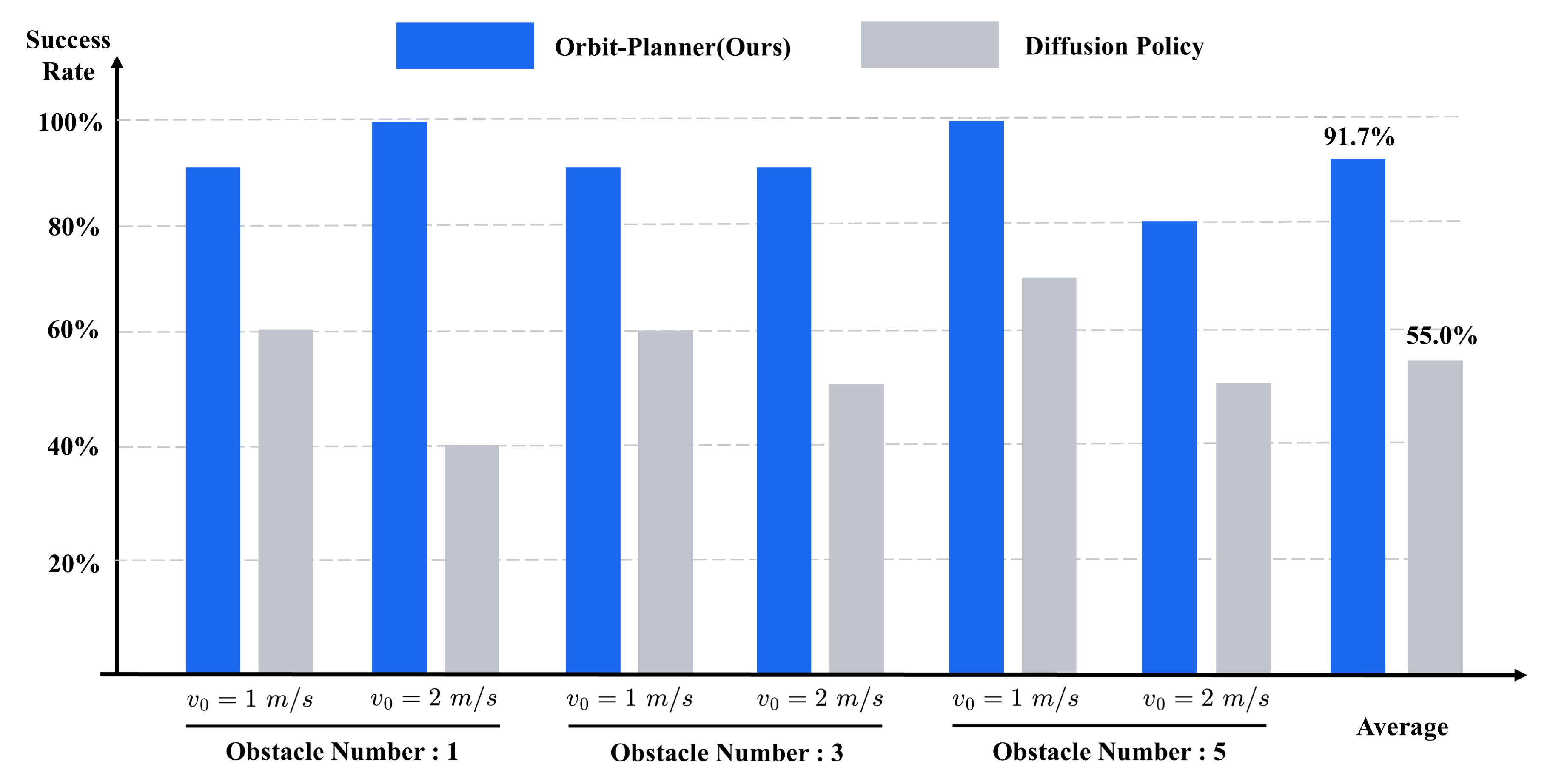}
	\figcaption{Closed-loop success rates of Orbit-Planner and Diffusion Policy~\cite{chi2025diffusion} in Isaac Sim under six settings (1/3/5 obstacles $\times$ $v_0=1/2$\,m/s) and their average.}
	\label{fig:nav_result}
\end{figure}

We further evaluate closed-loop obstacle-avoidance navigation in Isaac Sim. Orbit-Planner uses the frozen world model to imagine action-conditioned futures and select collision-free maneuvers, and is compared against an imitation-learning baseline, Diffusion Policy~\cite{chi2025diffusion}. Both methods are tested under six settings that vary the number of obstacles and the initial velocity, with 10 episodes per setting. In each episode, the obstacles are randomly placed and unseen during training. As shown in Fig.~\ref{fig:nav_result}, Orbit-Planner attains an average success rate of $91.7\%$, substantially outperforming Diffusion Policy ($55.0\%$). The advantage is consistent across all six settings and remains pronounced at higher speed and denser obstacle fields. These results indicate that action-conditioned latent rollouts enable anticipatory, physics-aware decisions, whereas a purely reactive imitation policy struggles to generalize when the scene dynamics deviate from the demonstrated distribution.

\section{Conclusion}
In this paper, we presented Orbit-Planner, a latent world model for on-orbit obstacle avoidance by autonomous satellite agents. Orbit-Planner learns action-conditioned spacecraft dynamics to perform future-state rollouts in latent space, employing a Physics Probe to decode physical state transitions from imagined trajectories. Experimental results demonstrate that the proposed model achieves long-horizon latent rollouts, physical-state readouts, and higher closed-loop navigation success than an imitation-learning baseline. In future work, we will study the sim-to-real gap under more realistic sensing and dynamics conditions.

\small
\bibliographystyle{IEEEtranN}
\bibliography{references}

\end{document}